\documentclass{article}

\usepackage{cite}
\usepackage{booktabs}
\usepackage{pifont}%
\usepackage{rotating}
\usepackage{amsfonts}
\usepackage{multirow}
\usepackage{color}
\usepackage{graphicx}
\usepackage{hyperref}
\usepackage{amsmath}
\usepackage{algorithm} %
\usepackage{flushend}
\usepackage{subcaption}
\usepackage{wrapfig}
\usepackage{breakurl}
\usepackage{enumitem}
\usepackage{xspace}
\usepackage[normalem]{ulem}
\usepackage{listings}
\usepackage{flushend}
\usepackage{footmisc}

\makeatletter
\newcommand{\@chapapp}{\relax}%
\makeatother
\usepackage[title,toc,titletoc,header]{appendix}

\usepackage{tikz}
\usetikzlibrary{positioning, shapes}

\usepackage[version=4]{mhchem} %
\usepackage{float} %

\usepackage{lipsum}

\newcommand{\DefMacro}[2]{\expandafter\newcommand\csname rmk-#1\endcsname{#2}}
\newcommand{\UseMacro}[1]{\csname rmk-#1\endcsname}

\newcommand{\XComment}[1]{}
\newcommand{\Space}[1]{}

\definecolor{gray}{RGB}{211,211,211}
\newcommand{\jbasicstyle}{\small\sffamily}

\newcommand{\jnumberstyle}{\scriptsize}

\lstdefinelanguage{pseudo}
{ morekeywords={for, in, break, continue, try, except, not,
  if,else,return,map,fieldElement_array_array40,fieldElement_array40},
  keywordstyle=\bfseries, lineskip=-0.1em, numbers=left,
  numberstyle=\jnumberstyle, numbersep=4pt, basicstyle=\jbasicstyle,
  breaklines=true, breakautoindent=true, tabsize=2,
  columns=fullflexible, morecomment=*[l][\textsl]{//},
  mathescape=true, }
\def\formulaname{Formula}

\newfloat{formula}{pb}{form}
\floatname{formula}{\formulaname}

\newcommand{\Relu}{ReLU\xspace}
\newcommand{\NNClass}{binary-weight ReLU\xspace}

\newcommand{\layerNum}{l}

\newcommand{\Section}[1]{\section{#1}}
\newcommand{\Subsection}[1]{\subsection{#1}}

\usepackage[disable,textwidth=60pt,textsize=scriptsize]{todonotes} %
\LetLtxMacro{\todom}{\todo}
\renewcommand{\todo}[1]{\todom[inline]{#1}}

 \usepackage[accepted]{icml2020}

\begin{document}

\icmltitlerunning{Deep Molecular Programming: A Natural Implementation of Binary-Weight ReLU Neural Networks}

\twocolumn[
  \icmltitle{Deep Molecular Programming:\\A Natural Implementation of Binary-Weight ReLU Neural Networks}

\icmlsetsymbol{equal}{*}

\begin{icmlauthorlist}
\icmlauthor{Marko Vasic}{ut}
\icmlauthor{Cameron Chalk}{ut}
\icmlauthor{Sarfraz Khurshid}{ut}
\icmlauthor{David Soloveichik}{ut}
\end{icmlauthorlist}

\icmlaffiliation{ut}{The University of Texas at Austin, USA}

\icmlcorrespondingauthor{Marko Vasic}{vasic@utexas.edu}

\icmlkeywords{Machine Learning, ICML}

\vskip 0.3in
]

\printAffiliationsAndNotice{}

\begin{abstract}
Embedding computation in molecular contexts incompatible with traditional electronics is expected to have wide ranging impact in synthetic biology, medicine, nanofabrication and other fields.
A key remaining challenge lies in developing programming paradigms for molecular computation that are well-aligned with the underlying chemical hardware and do not attempt to shoehorn ill-fitting electronics paradigms.
We discover a surprisingly tight connection between a popular class of neural networks (\NNClass aka BinaryConnect) and a class of coupled chemical reactions that are absolutely robust to reaction rates.
The robustness of rate-independent chemical computation makes it a promising target for bioengineering implementation.
We show how a BinaryConnect neural network trained in silico using well-founded deep learning optimization techniques, can be compiled to an equivalent chemical reaction network, providing a novel molecular programming paradigm.
We illustrate such translation on the paradigmatic IRIS and MNIST datasets.
Toward intended applications of chemical computation, we further use our method to generate a chemical reaction network that can discriminate between different virus types based on gene expression levels.
Our work sets the stage for rich knowledge transfer between neural network and molecular programming communities.
\end{abstract}

\Section{Introduction}

Although molecular computation cannot compete with electronics based on speed, the goal is to enable computation in contexts where traditional electronics cannot go.
Chemical control modules compatible with the chemical environments within natural or synthetic cells, bioreactors, and in-the-field diagnostics, are all envisioned applications for such technology.
Such computation could, for example, recognize disease state based on chemical inputs and actuate drug delivery to the affected cell.
The extensive information processing that occurs in natural cellular regulatory networks underlying all complex life, is a strong proof-of-principle that chemical computation is possible and useful.

Networks of coupled chemical reactions (chemical reaction networks, CRNs) are known to be Turing universal~\cite{soloveichik2008computation,fages2017strong}, but the resulting systematic ways of programming their behavior can result in extremely large reaction networks and exceedingly inefficient computation.
Reasoning in chemical reaction space is difficult: even a very small CRN can be hard to analyze.
Thus, there were many previous attempts to build intuitive, yet more efficient programming approaches for CRNs.
For example, \cite{senum2011rate} describe a number of computational modules including arithmetic modules and control flow.
Another example is CRN++~\cite{vasic2018crn}, an imperative programming language that compiles to CRNs.
Here, we present a different programming paradigm, one that allows translation of neural networks into chemical reactions.
In a similar way that writing an image classifier in an imperative language is prohibitively complex,
writing a classifier in a molecular programming language such as CRN++ is practically impossible.
However, the technique of programming using neural networks opens doors for such applications in the molecular programming community.

We focus on a class of neural networks called binary-weight \Relu neural networks.
These networks have binarized weights $\{+1, -1\}$, and were originally popularized due to their computational efficiency in electronics hardware~\cite{courbariaux2015binaryconnect}.
We show that \NNClass networks have a tight correspondence to a class of \emph{rate-independent} CRNs~\cite{chen2014deterministic,chen2014rate}.
In rate-independent CRNs,  computation arises solely from the stoichiometric exchange of reactants for products and the equilibrium is  independent of reaction rates.
The absolute robustness to reaction rates makes rate-independent CRNs a promising implementation target for bioengineering.

We first demonstrate our approach by training classifiers on the widely used machine learning datasets IRIS and MNIST.
A promising envisioned application of molecular computation lies in medical diagnostics and so called ``smart drugs'' which activate in response to specific molecular cues.
Thus, we next train a classifier to differentiate between four viral infections using chemical information as input (gene expression levels).
In all cases, we translate neural networks into CRNs, and simulate CRN behavior in a chemical kinetics simulation framework.
The chemical reaction networks exhibit the same output as their corresponding neural networks.

Section~\ref{sec:background} first reviews CRNs, and how we can view their behavior as computation, then reviews binary-weight \Relu networks and the BinaryConnect~\cite{courbariaux2015binaryconnect} method for training them.
In Section~\ref{sec:technique}, we first introduce a simple set of chemical reactions which compute the \Relu function, then describe our general technique to compile binary-weight \Relu networks into CRNs which compute the same output.
We also describe an optimization procedure which reduces the number of chemical reactions required to implement the neural network.
In Section~\ref{sec:experiments}, we give simulation results on our chemical classifiers for IRIS, MNIST, and viral infection classification, and verify that their outputs match the neural networks they implement.

\Section{Background}
\label{sec:background}
In this section we provide a brief description of chemical reaction networks and BinaryConnect neural networks.

\Subsection{Chemical Reaction Networks}
Chemical reaction networks (CRNs) formally model the time evolution of concentrations of chemical species in a solution undergoing chemical change.
CRNs are typically used to understand naturally occurring chemistry and abstract the behavior of existing biological regulatory networks.
The CRN formalism thus provides a standardized way to specify the interaction rules that we expect interacting chemicals to obey in order to perform computation.
Closely related models from distributed computing include population protocols~\cite{angluin2006computation}, Petri nets~\cite{petri1966communication}, and vector addition systems~\cite{karp1969parallel}.

A CRN is formally a set of chemical \emph{species} (typically written with capital letters) and a set of \emph{reactions} between them.
A \emph{state} of the CRN is an assignment of a nonnegative real value to each species, representing the species' concentrations (quantity per unit volume).
As the state varies over time, we denote by $a(t)$ the concentration of species $A$ at time $t$.
Each reaction is composed of two multisets over the species, the \emph{reactants} and \emph{products}, which specify the stoichiometry (the relative quantities of the involved species).
For example, species $A$ and $B$ might react to produce species $C$,
which we write as the reaction
$$A+B \longrightarrow C.$$
Each reaction has an associated \emph{rate constant} which affects the dynamics of state change of the system.
Under typical mass-action kinetics,
the instantaneous rate of a reaction is the product of the rate constant and the concentrations of the reactants (i.e., $k \cdot a(t) \cdot b(t)$ for the reaction above).\footnote{Although the details of CRN kinetics is not essential for understanding the results in this paper, the numerical simulations are done using continuous mass-action kinetics.
In other models of chemical kinetics the amounts of species are represented by non-negative integer molecular counts and the system corresponds to a continuous time Markov chain. Our results apply to this setting as well, as long as real-valued signals are properly discretized.}
The behavior of the whole system is then described by a set of ordinary differential equations (ODEs) generated based on the reactions.
For example, if the reaction above is the only one producing $C$, the kinetics of $C$ is governed by $dc(t)/dt = k \cdot a(t) \cdot b(t)$.

To define CRN computation, some chemical species are considered \emph{input species}, and some others, disjoint from the first, are considered \emph{output species}.
The initial concentrations of the input species are the input to the computation.
The system asymptotically converges to a state where the concentrations of the output species are equal to the desired output values.

Many CRN computations work under strong assumptions about the rates of the reactions, but engineering reactions with precise rates is difficult, and small changes in the environment can further disturb these rates.
Instead, computation can be achieved by stoichiometry alone~\cite{chen2014deterministic,chen2014rate}, which forms the basis of the computation considered in this paper.
The essential example is the reaction we have already seen: $A + B \longrightarrow C$.
The concentration of $C$ as time approaches infinity is the $\min$ of the initial concentrations of $A$ and $B$, since the reaction can occur only as long as both $A$ and $B$ are present.
This fact holds for any (nonzero) reaction rate.
Thus we say this CRN rate-independently computes the $\min$ function: $c = \min(a(0), b(0))$.

A programmable chemical process called DNA strand displacement can in principle (and, to some extent, experimentally~\cite{chen2013programmable,srinivas2017enzyme}) implement arbitrary, rationally designed CRNs~\cite{soloveichik2010dna}.
Watson-Crick base pairing enables programmability, since interactions are mediated by nucleotide sequence complementarity.
In the basic DNA strand displacement reaction, an ``output'' strand is initially bound to a complementary strand.
A single-stranded ``input'' binds to the double-stranded complex, displacing the ``output'' strand.
Extended designs based on this elementary reaction allow enzyme-free synthetic DNA systems to behave as arbitrary CRNs.\footnote{No reaction is truly irreversible; that is to say, if $A +B \longrightarrow C$, then $C \longrightarrow A + B$ with some non-zero rate.
However, nature provides many examples of reactions which are effectively irreversible due to a large change in free energy between reactions and products.
Reactions implemented by DNA strand displacement are thermodynamically driven by the formation of additional nucleotide base pairs,
which provide enough free energy for the reaction to behave effectively irreversibly.}

Rate-independent computation aligns well with the design of DNA strand displacement systems.
The stoichiometry of a strand displacement reaction is engineered by setting regions on two or more strands to be complementary or orthogonal.
In contrast, setting precise reaction rates requires designing specific sequences that achieve desired hybridization free energies.
Hybridization rates are highly dependent on the temperature and on the chemical composition of the solution.
Further, for a particular chemical implementation, there is some maximum rate possible, so requiring some reactions to be fast and others slow necessarily requires slowing the entire system down.

Although concentrations of species are nonnegative, to imitate the computation done by a neural network we need to store and process negative values.
To do so, we represent negative values by the \emph{dual-rail} convention: a value $x$ is represented not by the concentration of one species $X$, but by the difference in concentration between two species $X^+$ and $X^-$~\cite{chen2014rate}. More precisely, at time $t$, $x(t) = x^+(t) - x^-(t)$.
Note that values have multiple representations:
$x = 3$ can be represented as either $x^+ = 3$ and $x^- = 0$ or $x^+ = 4$ and $x^- = 1$.
Although the value is the same, the choice of representation affects convergence time.
(Convergence time is an important question for future work as mentioned in Section~\ref{sec:conclusion}.)

\Subsection{Binary-Weight \Relu (BinaryConnect)}
\label{sec:binaryconnect}
In recent years we have seen a growing interest in building specialized deep learning hardware to enable new and more complex deep learning applications as well as deploying deep learning systems on low-power devices~\cite{dean2012large,kim2009highly,chen2014diannao}.
One direction of work tries to tackle this problem by designing more computationally efficient neural networks~\cite{courbariaux2015binaryconnect,hubara2016binarized,li2016ternary,simons2019review}.
One of the initial works in this area introduces BinaryConnect networks~\cite{courbariaux2015binaryconnect}
which restrict the weights of neural networks to values $\pm 1$.
This restriction on the weight values enables replacement of most of the multiply units by simple accumulators which leads to a drastic reduction of computational expenses,
considering that the multipliers are the most space and power hungry components of specialized deep learning hardware.
Moreover, in spite of severe restriction on the weight values, BinaryConnect still achieves near state-of-the-art results on standard machine learning datasets.

BinaryConnect networks are trained similarly to traditional neural networks through a repeated sequence of:
(a) forward-propagation pass computing the output of a network given an input;
(b) backward-propagation pass computing error terms (derivatives of the cost function in respect to weights and bias terms);
and (c) parameter updates (updating weight and bias terms based on the error terms).
BinaryConnect maintains real-valued weights throughout training, but discretizes them to $1$ or $-1$ (if positive or negative, respectively) during the forward and backward-propagation passes.
The parameter update is then performed on the real-valued weights.
The final binary-weight neural network is generated by discretizing the resulting real-valued weights at the end of training.
(Other variations of BinaryConnect training, such as stochastic discretization, are described in the original paper.)

\Section{Technique}
\label{sec:technique}

First we describe the technique for compiling \NNClass networks to CRNs.
Then we show how to optimize the CRNs to reduce the total number of reactions.
Ultimately, there will be one reaction per \Relu node.

\Subsection{\Relu Chemical Reaction Network}
We propose the following CRN to compute the \Relu function:
\begin{align}
  \ce{$X^+$ &->[] $M$ + $Y^+$} \label{eq:relu1} \\
  \ce{$M$ + $X^-$ &->[] $Y^-$} \label{eq:relu2}
\end{align}
In order to understand why the above reactions compute \Relu,
suppose we start with $x^+(0)$ amount of $X^+$ and $x^-(0)$ amount of $X^-$ and no other species.
These concentrations represent the input value $x(0) = x^+(0) - x^-(0)$ in dual-rail form.
Although the two reactions will be happening in parallel, the first reaction converges to producing $x^+(0)$ amount of $M$ and $Y^+$.
Therefore, the second reaction converges to producing $\min(x^+(0), x^-(0))$ of $Y^-$.
This implies that the system converges to the dual-rail value of the output $y = y^+(t) - y^-(t)  =  x^+(0) - \min(x^+(0), x^-(0)) = (x^+(0) - x^-(0)) - \min(x^+(0) - x^-(0), x^-(0) - x^-(0)) = x - \min(x,0) = \max(x,0)$.
Note that this CRN is rate-independent because the computation will be correct no matter what the rates of the individual reactions are.

Since the output species $Y^+$ and $Y^-$ are not consumed in the above CRN,
they can be used as input species for any composed downstream computation without interfering with the computation of \Relu.
Reactions \ref{eq:relu1}--\ref{eq:relu2} form the simplest CRN computing \Relu in the sense that there is no composable CRN with fewer than $2$ reactions and $5$ species
computing this function~\cite{vasic2019crns}.\footnote{There is a simpler, non-composable \Relu computing CRN: $X^+ \longrightarrow Y^+$, $Y^+ + X^- \longrightarrow W$. ($W$ is an inert waste species.)
It cannot be combined with another CRN which uses species $Y^+$ as an input, as $Y^+$ may be consumed by the downstream CRN before it is annihilated by the second reaction.
Composability is a well-understood feature in rate-independent CRNs~\cite{ChalkKRS18,severson19}.}

\Subsection{Compiling Binary-Weight \Relu Networks}
\label{sec:compiling-binconnect}

\begin{figure}[!t]
  \centering
  \includegraphics[scale=0.65]{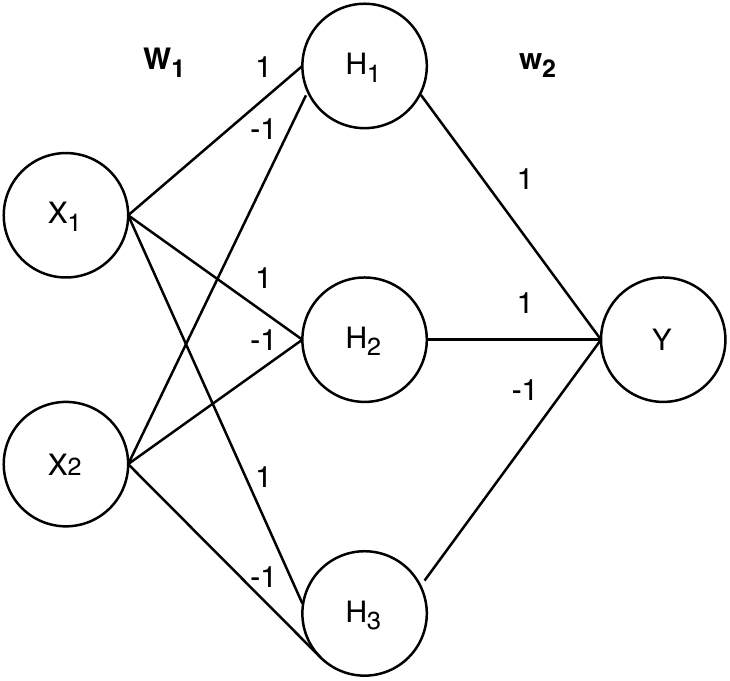}
  \caption{Binary-weight neural network.}
  \label{fig:nn}
\end{figure}

Figure~\ref{fig:nn} shows an example \NNClass network.
This network consists of an input layer, a single hidden layer with the \Relu activation function, and an output layer without an activation function.
Let us initially ignore the bias terms.
Then the output of the network is defined by $ y = \Relu(\boldsymbol{x}^\top \cdot \boldsymbol{W_1}) \cdot \boldsymbol{w_2} $,
where $\boldsymbol{x} \in \mathbb{R}^2$ is an input vector, $\boldsymbol{W_1} \in \{-1, 1\}^{2 \times 3}$ is a weight matrix into the hidden layer, $\boldsymbol{w_2} \in \{-1, 1\}^3$ is a weight vector into the output layer, and $y \in \mathbb{R}$ is the output.
The Figure~\ref{fig:nn} neural network weight values are:
\[
\boldsymbol{W_1}=
  \begin{bmatrix}
    1  &  1 & 1 \\
    -1 & -1 & -1
  \end{bmatrix},
\boldsymbol{w_2}=
  \begin{bmatrix}
    1 & 1 & -1
  \end{bmatrix}^\top
\]
We now convert this \NNClass network into a CRN.
Initially, for each input $x_i$ in the input vector, species $X_i^+$ and $X_i^-$ will be present such that $x_i^+(0) - x_i^-(0) = x_i$.
We multiply the input vector by the weights $\boldsymbol{W_1}$ by the following reactions:
\begingroup
\allowdisplaybreaks
\begin{align}
  \ce{$X_1^+$ &->[] $I_{1,1}^+$ + $I_{1,2}^+$ + $I_{1,3}^+$} \label{crn:first} \\
  \ce{$X_1^-$ &->[] $I_{1,1}^-$ + $I_{1,2}^-$ + $I_{1,3}^-$} \\
  \ce{$X_2^+$ &->[] $I_{1,1}^-$ + $I_{1,2}^-$ + $I_{1,3}^-$} \\
  \ce{$X_2^-$ &->[] $I_{1,1}^+$ + $I_{1,2}^+$ + $I_{1,3}^+$}
\end{align}
\endgroup
Species $I_{l,i}$ represents the $i$th intermediate (before applying nonlinearity) species of the layer $l$.
For a weight with value $1$ we include a reaction with a positive input species as a reactant and positive output species as a product, as that has the effect of addition to the product species' value.
For a weight with value $-1$ we include a reaction with a positive input species as a reactant and negative output species as a product, as that has the effect of subtraction from the product species' value.
To maintain the dual-rail convention we also include a reaction which contains all the same species with signs flipped.
Next, to implement the \Relu activation function we use the module discussed above (reactions \ref{eq:relu1}--\ref{eq:relu2}), with the appropriate renaming of species:
\begin{align}
  \ce{$I_{1,1}^+$ &->[] $M_{1,1}$ + $H_{1,1}^+$} \\
  \ce{$M_{1,1}$ + $I_{1,1}^-$ &->[] $H_{1,1}^-$}
\end{align}%
\begin{align}
  \ce{$I_{1,2}^+$ &->[] $M_{1,2}$ + $H_{1,2}^+$} \\
  \ce{$M_{1,2}$ + $I_{1,2}^-$ &->[] $H_{1,2}^-$}
\end{align}%
\begin{align}
  \ce{$I_{1,3}^+$ &->[] $M_{1,3}$ + $H_{1,3}^+$} \\
  \ce{$M_{1,3}$ + $I_{1,3}^-$ &->[] $H_{1,3}^-$}
\end{align}
Each intermediate species $I_{l,i}$ of the hidden layer $l$ is a part of a separate \Relu module producing output species $H_{l,i}$.
Finally, to multiply by the weight vector $\boldsymbol{w_2}$ we use the following reactions (similar to the multiplication by $\boldsymbol{W_1}$):
\begingroup
\allowdisplaybreaks
\begin{align}
  \ce{$H_{1,1}^+$ &->[] $Y^+$} \label{crn:downstream-uni-first} \\
  \ce{$H_{1,1}^-$ &->[] $Y^-$} \\
  \ce{$H_{1,2}^+$ &->[] $Y^+$} \\
  \ce{$H_{1,2}^-$ &->[] $Y^-$} \\
  \ce{$H_{1,3}^+$ &->[] $Y^-$} \\
  \ce{$H_{1,3}^-$ &->[] $Y^+$} \label{crn:last}
\end{align}
\endgroup
The $Y$ species are the output species, and the value $y = y^+(t) - y^-(t)$ approaches $\Relu(\boldsymbol{x}^\top \cdot \boldsymbol{W_1}) \cdot \boldsymbol{w_2}$ as $t$ goes to infinity.

The complete equation of the neural network with bias terms is: $ y = \Relu(\boldsymbol{x}^\top \cdot \boldsymbol{W_1} + \boldsymbol{b_1}^\top) \cdot \boldsymbol{w_2} + b_2$.
To encode bias terms $\boldsymbol{b_1}$, it is enough to set initial concentrations of the hidden layer $I_{1,i}$ species accordingly (e.g., ${i_{11}^+(0) - i_{11}^-(0)}$ should be equal to the first component of vector $\boldsymbol{b_1}$); similarly for $b_2$ and $Y$ species.

\begin{algorithm}[tb]
  \begin{small}
   \caption{NNCompile($\pm1$-weight neural network: \textit{nn})}
   \label{alg:nnCompiler}
\begin{algorithmic}[1]
  \STATE $crn = newCRN()$
  \FOR{$\layerNum=2$ {\bfseries to} $nn.numLayers()$}
    \STATE $layer = nn.getLayer(\layerNum)$
    \STATE $W, b, a = layer.params()$
    \STATE $pName = \text{`}H\text{'}$ {\bfseries{if}} $a.linear()$ {\bfseries{else}} $\text{`}I\text{'}$ \label{alg:nnCompiler:line:pName}
    \FOR{$i = 1$ {\bfseries to} $W.dimsX$}
      \STATE $rxn = newReaction()$
      \STATE $rxn.reactants = \{ X_{i}^+$ {\bfseries{if}} $l == 1$ {\bfseries{else}} $H_{\layerNum-1,i}^+ \}$ \label{alg:nnCompiler:line:reactants}
      \FOR{$j = 1$ {\bfseries to} $W.dimsY$}
        \STATE $pSign = \text{`}+\text{'}$ {\bfseries{if}} $W[i,j] == 1$ {\bfseries{else}} $\text{`}-\text{'}$ \label{alg:nnCompiler:line:pSign}
        \STATE $rxn.products.add(pName_{\layerNum,j}^{pSign})$ \label{alg:nnCompiler:line:productsAdd}
      \ENDFOR
      \STATE $crn.add(rxn)$ \label{alg:nnCompiler:line:add1}
      \STATE $crn.add(rxn.reverseSigns())$ \label{alg:nnCompiler:line:add2}
    \ENDFOR
    \FOR{$j = 1$ {\bfseries to} $W.dimsY$}
      \IF{$b[j] > 0$} \label{alg:nnCompiler:line:setConcStart}
        \STATE $crn.setConc(pName_{\layerNum,j}^{+}, b[j])$
      \ELSE
        \STATE $crn.setConc(pName_{\layerNum,j}^{-}, -b[j])$
      \ENDIF \label{alg:nnCompiler:line:setConcEnd}
      \IF{$a.nonlinear()$}
        \STATE $crn.add(I_{\layerNum,j}^+ \longrightarrow M_{\layerNum,j} + H_{\layerNum,j}^+ )$
        \STATE $crn.add(M_{\layerNum,j} + I_{\layerNum,j}^- \longrightarrow H_{\layerNum,j}^- )$
      \ENDIF
    \ENDFOR
  \ENDFOR
  \RETURN $crn$
\end{algorithmic}
\end{small}
\end{algorithm}

We summarize the compilation procedure in Algorithm~\ref{alg:nnCompiler}.
The input of the algorithm is a \NNClass network ($nn$), and the output is its CRN equivalent.
We iterate through all the layers of $nn$ starting from the first hidden layer and including the output layer (layer number $1$ is the input layer).
If a layer does not contain the nonlinear activation (\Relu) we include reactions directly producing the layer output species (named $H$), otherwise intermediate species are produced first (named $I$); the name of the appropriate product species is stored in $pName$ (line~\ref{alg:nnCompiler:line:pName}).
We iterate through the rows of the weight matrix ($W$), creating one reaction per row, where the reactant is $H_{\layerNum-1,i}^+$ ($i$th unit of the previous layer), or $X_{i}^+$ in the case where the previous layer is the input layer (line~\ref{alg:nnCompiler:line:reactants}).
To construct the products of the reaction, we iterate through all of the columns of $W$, adding $pName$ species with positive sign when the weight is $+1$ or negative sign when the weight is $-1$ (lines~\ref{alg:nnCompiler:line:pSign}--\ref{alg:nnCompiler:line:productsAdd}).
For a valid dual-rail computation, we include such a constructed reaction to the CRN (line~\ref{alg:nnCompiler:line:add1}), as well as the same reaction with the signs of all the species flipped (line~\ref{alg:nnCompiler:line:add2}).
Next, we set the initial concentrations of $pName$ species to correspond to the bias terms.
If a bias term is positive, we set the positive species to $b[j]$, otherwise we set the negative species to $-b[j]$ (lines~\ref{alg:nnCompiler:line:setConcStart}--\ref{alg:nnCompiler:line:setConcEnd}).
Finally, if the layer contains a \Relu activation, we include the \Relu CRN taking as inputs layer intermediate ($I$) species and producing layer output ($H$) species.
Note that unlike in the previous example, the network output species will be denoted by $H_{N,j}$ (where $N$ is the number of layers in the network).

To compute with the CRN, we provide input $\boldsymbol{x}$ by setting the initial concentrations of the input species $X$ as follows.
If a component $x_i$ of the input vector is non-negative, then the concentration of the positive input species $X_i^{+}$ is set to $x_i$;
otherwise the concentration of the negative input species $X_i^{-}$ is set to $-x_i$.
Note that in chemistry, reactions occur in parallel, so layers begin their computation before the previous ones complete, unlike typical neural network architectures.
\todom{it sounds a bit weird to put this here about reactions occurring in parallel}

\Subsection{Optimization: Reducing the Compiled CRN}\label{subsec:optimization}

We find that unimolecular reactions, such as the first reactions of \Relu modules, can be eliminated from the CRN by altering the bimolecular reactions and the initial concentrations of the CRN species, a process which we describe next.
Unimolecular reactions are those with exactly one reactant like $A \longrightarrow B + C$.
Whenever $A$ is produced in another reaction, we can replace it with $B + C$.
For example, if there is another reaction $X \longrightarrow A + B$, we replace the reaction with $X \longrightarrow 2B+C$.
Further, we adjust the initial concentrations of the product species ($B$ and $C$) by increasing them by the initial concentrations of the reactant ($A$).
Importantly, this transformation works only if $A$ is not a reactant in any other reaction;
for example, if there were another reaction like $X + A \longrightarrow Y$, it is not clear what to replace instances of $A$ with, and indeed it is not possible to remove the unimolecular reaction in that case.
Luckily, our construction has the property that any species occurs as a reactant in at most one reaction.

We will now illustrate the aforementioned elimination procedure on the neural network CRN presented in Section~\ref{sec:compiling-binconnect}, characterized by reactions~\ref{crn:first}--\ref{crn:last}.
We start by removing the downstream-most unimolecular reactions (reactions~\ref{crn:downstream-uni-first}--\ref{crn:last}), although the order does not matter.
Removing these reactions and adjusting the products accordingly results in the following CRN:
\begingroup
\allowdisplaybreaks
\begin{align}
  \ce{$X_1^+$ &->[] $I_{1,1}^+$ + $I_{1,2}^+$ + $I_{1,3}^+$} \\
  \ce{$X_1^-$ &->[] $I_{1,1}^-$ + $I_{1,2}^-$ + $I_{1,3}^-$} \\
  \ce{$X_2^+$ &->[] $I_{1,1}^-$ + $I_{1,2}^-$ + $I_{1,3}^-$} \\
  \ce{$X_2^-$ &->[] $I_{1,1}^+$ + $I_{1,2}^+$ + $I_{1,3}^+$} \\
  \ce{$I_{1,1}^+$ &->[] $M_{1,1}$ + $Y^+$} \label{crn:uni-1} \\
  \ce{$M_{1,1}$ + $I_{1,1}^-$ &->[] $Y^-$} \\
  \ce{$I_{1,2}^+$ &->[] $M_{1,2}$ + $Y^+$} \label{crn:uni-2} \\
  \ce{$M_{1,2}$ + $I_{1,2}^-$ &->[] $Y^-$} \\
  \ce{$I_{1,3}^+$ &->[] $M_{1,3}$ + $Y^-$} \label{crn:uni-3} \\
  \ce{$M_{1,3}$ + $I_{1,3}^-$ &->[] $Y^+$}
\end{align}
\endgroup
Note that the initial concentrations of $Y$ species are unaffected as species $H_{1,i}$ are initially set to $0$.
Next, we remove the now downstream-most unimolecular reactions (reactions~\ref{crn:uni-1},\ref{crn:uni-2},\ref{crn:uni-3}) and obtain the following CRN:
\begingroup
\allowdisplaybreaks
\begin{align}
  \ce{$X_1^+$ ->[] $M_{1,1}$ &+ $M_{1,2}$ + $M_{1,3}$ + $2Y^+$ + $Y^-$} \label{crn:uni-2-start} \\
  \ce{$X_1^-$ &->[] $I_{1,1}^-$ + $I_{1,2}^-$ + $I_{1,3}^-$} \\
  \ce{$X_2^+$ &->[] $I_{1,1}^-$ + $I_{1,2}^-$ + $I_{1,3}^-$} \\
  \ce{$X_2^-$ ->[] $M_{1,1}$ &+ $M_{1,2}$ + $M_{1,3}$ + $2Y^+$ + $Y^-$} \label{crn:uni-2-end} \\
  \ce{$M_{1,1}$ + $I_{1,1}^-$ &->[] $Y^-$} \\
  \ce{$M_{1,2}$ + $I_{1,2}^-$ &->[] $Y^-$} \\
  \ce{$M_{1,3}$ + $I_{1,3}^-$ &->[] $Y^+$}
\end{align}
\endgroup
The initial concentrations of $M_{1,1}$, $M_{1,2}$, $M_{1,3}$, $Y^+$ and $Y^-$ are also affected:
\begin{align*}
  m_{1,i}(0) &\mathrel{+}= i_{1,i}^+(0), \forall i \in \{1, 2, 3\} \\
  y^+(0) &\mathrel{+}= i_{1,1}^+(0) + i_{1,2}^+(0) \\
  y^-(0) &\mathrel{+}= i_{1,3}^-(0)
\end{align*}
The only remaining unimolecular reactions are those whose reactants are input species ($X$).
We do not apply the above elimination procedure to these unimolecular reactions because doing so would remove the input species entirely.
Then giving the input to the CRN would require setting the initial concentrations of a number of other species,
complicating the input representation.

Note that the CRN can be further simplified while preserving dual-rail values.
Reactions \ref{crn:uni-2-start} and \ref{crn:uni-2-end} can be reduced to:
\begin{align}
  \ce{$X_1^+$ ->[] $M_{1,1}$ &+ $M_{1,2}$ + $M_{1,3}$ + $Y^+$} \\
  \ce{$X_2^-$ ->[] $M_{1,1}$ &+ $M_{1,2}$ + $M_{1,3}$ + $Y^+$}
\end{align}
Further, the initial concentrations of $Y^+$ and $Y^-$ can be reduced by the same amount to make the smaller one zero.

In general, the CRN resulting from the optimization procedure has the following properties.
There are no unimolecular reactions besides the input layer, for which there are two reactions per input.
There is one bimolecular reaction per ReLU node of the original network.
There are no other reactions.
Optimization of some adversarial ReLU networks results in reactions with a number of products exponential in the depth of the network.
Understanding the scaling of the number of products is an important avenue for future work to ensure feasible CRNs.

\begin{algorithm}[tb]
  \begin{small}
   \caption{reduce(CRN: $crn$)}
   \label{alg:crnReduce}
   \begin{algorithmic}[1]
     \STATE $uniRxns = crn.getUnimolecularReactions()$ \label{alg:crnReduce:line:getRxns}
     \FORALL{$uniRxn \in uniRxns$}
       \STATE $R = uniRxn.getFirstReactant()$
       \STATE {\bfseries{if}} $crn.inputSpecies(R)$ {\bfseries{then \textit{continue}}} \label{alg:crnReduce:line:continue}
       \STATE $r(0) = crn.getConc(R)$
       \FORALL{$rxn_i \in crn.reactions$} \label{alg:crnReduce:line:alterStart}
         \IF{$rxn_i.containsProduct(R)$}
           \STATE $rxn_i.removeProduct(R)$
           \STATE $rxn_i.addProducts(uniRxn.products)$
         \ENDIF
       \ENDFOR \label{alg:crnReduce:line:alterEnd}
       \FORALL{$P_i \in uniRxn.products$}
       \STATE $crn.addConc(P_i, r(0))$ \label{alg:crnReduce:line:addConc}
       \ENDFOR
       \STATE $crn.remove(uniRxn)$ \label{alg:crnReduce:line:remove}
       \ENDFOR
       \FORALL{$S^+, S^- \in crn.getDualRailPairs()$}
         \FORALL{$rxn_i \in crn.reactions$} \label{alg:crnReduce:line:simplifyProductsStart}
           \WHILE{$S^+, S^- \in rxn_i.products$}
             \STATE $rxn.removeProduct(S^+)$
             \STATE $rxn.removeProduct(S^-)$
           \ENDWHILE
         \ENDFOR \label{alg:crnReduce:line:simplifyProductsEnd}
         \STATE $m = min(crn.getConc(S^+), crn.getConc(S^-))$ \label{alg:crnReduce:line:simplifyStart}
         \STATE $crn.addConc(S^+, -m)$
         \STATE $crn.addConc(S^-, -m)$ \label{alg:crnReduce:line:simplifyEnd}
       \ENDFOR
   \end{algorithmic}
   \end{small}
\end{algorithm}

Algorithm~\ref{alg:crnReduce} summarizes the reduction procedure.
The algorithm first collects a list of all unimolecular reactions (line~\ref{alg:crnReduce:line:getRxns}).
Then, for each unimolecular reaction ($R \longrightarrow \sum_i{P_i}$) the following steps are done:
(a) if reactant $R$ is an input species the following steps are skipped and the next reaction is processed (line~\ref{alg:crnReduce:line:continue});
(b) for every reaction in the CRN producing $R$, the list of products is altered by removing the product $R$ and adding products $P_i$ (lines~\ref{alg:crnReduce:line:alterStart}--\ref{alg:crnReduce:line:alterEnd});
(c) the initial concentrations of species $P_i$ are increased by the amount of the initial concentration of $R$ (line~\ref{alg:crnReduce:line:addConc}); and
(d) the unimolecular reaction is removed from the CRN (line~\ref{alg:crnReduce:line:remove}).
After these steps we perform the following behavior-preserving transformations that further simplify the CRN.
For each dual-rail value represented by $S^+$ and $S^-$ species we:
(i) for every reaction whose products multiset contain both $S^+$ and $S^-$,
remove both species as long as both are present
(lines~\ref{alg:crnReduce:line:simplifyProductsStart}--\ref{alg:crnReduce:line:simplifyProductsEnd});
(ii) reduce the initial concentrations of $S^+$ and $S^-$ by the minimum of their values (lines~\ref{alg:crnReduce:line:simplifyStart}--\ref{alg:crnReduce:line:simplifyEnd}).

\Section{Experiments}
\label{sec:experiments}
In this section we describe computational experiments showcasing compilation from neural networks to CRNs.
We train BinaryConnect networks on IRIS~\cite{fisher1936use,anderson1936species}, MNIST~\cite{lecun1998gradient}, and virus infection datasets~\cite{GSE73072}.
We then translate trained neural networks to CRNs following our compilation technique (Section~\ref{sec:technique}),
and simulate the reactions' behavior using an ODE simulator~\cite{CRNSimulatorPackage}.
The simulator models CRNs with real-valued concentrations in the standard mass-action model of chemical kinetics, where the rate of a reaction is directly proportional to the concentrations of the reactants.
Our main goal is to show the equivalence of the trained neural network and the compiled CRN, and not to improve accuracy of machine learning models, which is orthogonal to our work.

\Subsection{Datasets}

\textbf{IRIS}. The IRIS dataset consists of $150$ examples of $3$ classes of flowers (Setosa, Versicolor or Virginica), and $4$ features per example (sepal length and width, and petal length and width).
Considering the small dataset size and that our goal is to show the equivalence of the neural network and the translated CRN,
we use the whole dataset for both training and evaluation.

\textbf{MNIST}. The MNIST dataset consists of labeled handwritten digits, where features are image pixels, and labels are digits ($0$ to $9$).
We split the original MNIST training set consisting of $60,000$ images into $50,000$ for the training set, and $10,000$ for the validation set.
We use the original test set consisting of $10,000$ images.
In preprocessing, we center the images (as is done in the BinaryConnect work), and additionally we scale them from $28 \times 28$ to $14 \times 14$.

\textbf{Virus Infection}. For the virus infection classifier, we used data from NCBI GSE73072~\cite{GSE73072}.
The dataset contains microarray data capturing gene expression profiles of humans, with the goal of studying four viral infections: H1N1, H3N2, RSV, and HRV (labels).
There are $148$ patients in the dataset, each with about $20$ separate profiles taken at different times during their infection period, for a total of $2,886$ samples.
The dataset contains information about which patient was infected and during which point of time.
We filter the samples leaving only those that correspond to an active infection, and thus make the data suitable for classification of the four viruses.
Finally, we have a total of $698$ examples, split into $558$ for training, $34$ for validation, and $104$ for testing.
Each sample measures expression of $12,023$ different genes (features); we use the $10$ most relevant genes as features which are selected using GEO2R tool~\cite{GEO2R} from the NCBI GEO.

\Subsection{Results}

\begin{figure*}[!t]
  \centering
  \begin{subfigure}[!t]{0.48\textwidth}
    \centering
    \includegraphics[scale=0.52]{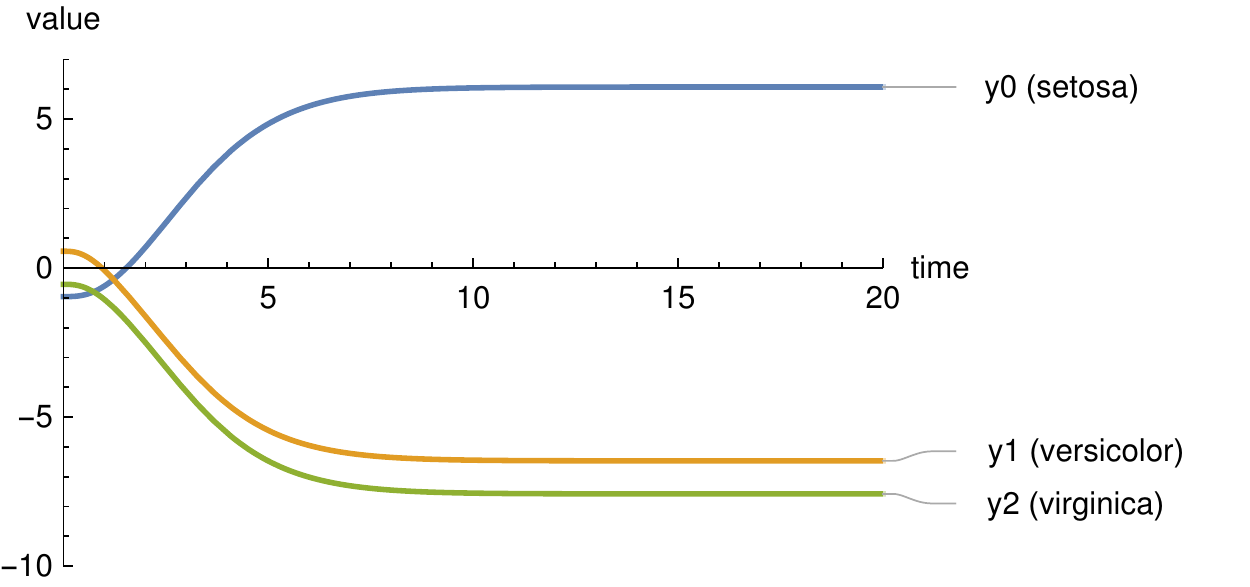}
    \caption{IRIS CRN Simulation.}
  \end{subfigure}
  \begin{subfigure}[!t]{0.48\textwidth}
    \centering
    \includegraphics[scale=0.52]{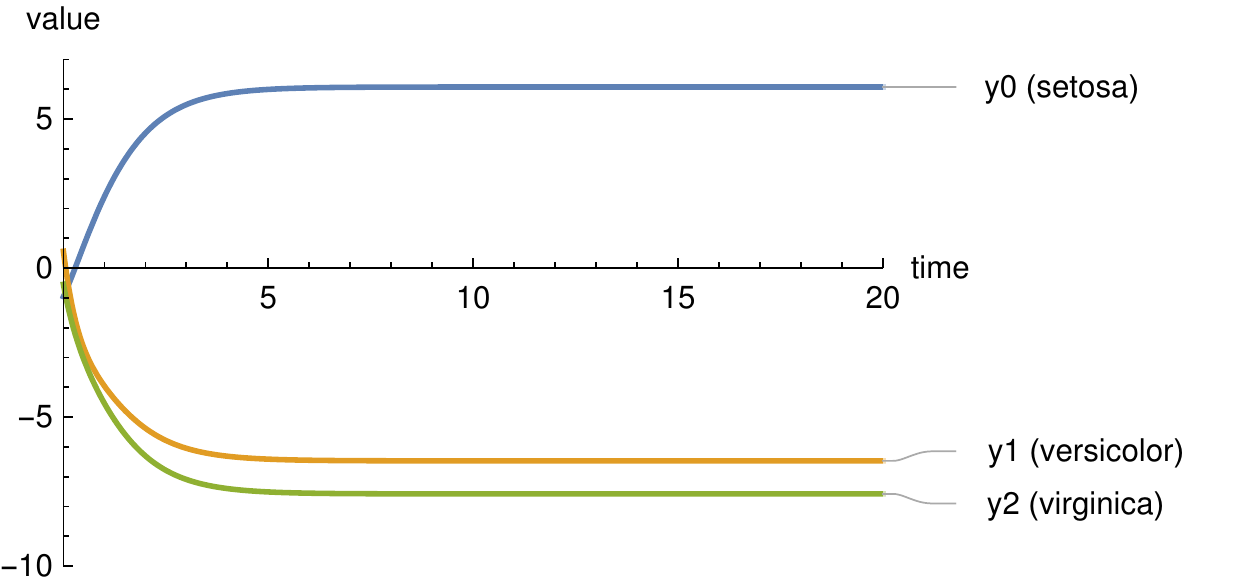}
    \caption{IRIS Reduced CRN Simulation.}
  \end{subfigure}
  \caption{Simulation of IRIS CRN (left) and IRIS CRN in reduced form (right) on the same example. Plots show concentrations of the output species over time, i.e., the difference between positive and negative output species ($y_i(t) = y_i^+(t) - y_i^-(t)$). Networks classify given input with label $0$.
  The reduced CRN converges significantly faster.
  Note that the initial concentrations of output species change after optimization based on the biases in the ReLU network, as described in Section~\ref{subsec:optimization}.}
  \label{fig:iris-simulations}
\end{figure*}

\textbf{IRIS}. We train a neural network with a single hidden layer consisting of $8$ units, $4$ input units (capturing the features of IRIS flowers), and $3$ output units where the unit with the highest value determines the output class.
Given that our primary goal is to show the equivalence of a neural network and the compiled CRN, and since the dataset size is quite small ($150$ examples), we train and evaluate on the whole IRIS dataset.
We achieve accuracy of $94$\% ($141$ out of $150$ examples correctly classified) with a trained BinaryConnect neural network.
The equivalent CRN consists of $40$ chemical reactions (unoptimized compilation), and $16$ chemical reactions (optimized compilation).
We simulate both versions of the CRN and confirm that their outputs (labels) match the neural network in all $150$ examples.
Figure~\ref{fig:iris-simulations} shows the simulation results of the IRIS CRNs on an example input,
which is classified with label $0$ (since the $y_0$ output is largest).
In both CRNs, the final simulation values of $y_0$, $y_1$ and $y_2$ match the output units values of the neural network (approaching up to $4$ decimal places in the time of simulation).
Thus not only does the CRN have the same classification output,
but also the numerical values of the individual dual-rail outputs match the output units of the neural network.
In regard to the dynamics of CRN convergence, the reduced CRN converges significantly faster.

\begin{figure*}[!t]
  \centering
  \begin{subfigure}[!t]{0.48\textwidth}
    \centering
    \includegraphics[scale=0.4]{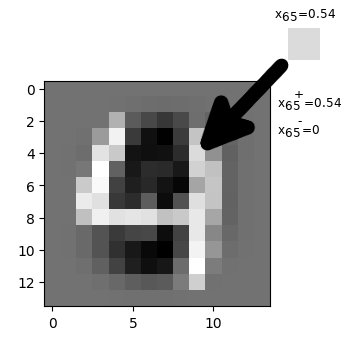}
    \caption{Input image.}
  \end{subfigure}
  \begin{subfigure}[!t]{0.48\textwidth}
    \centering
    \includegraphics[scale=0.5]{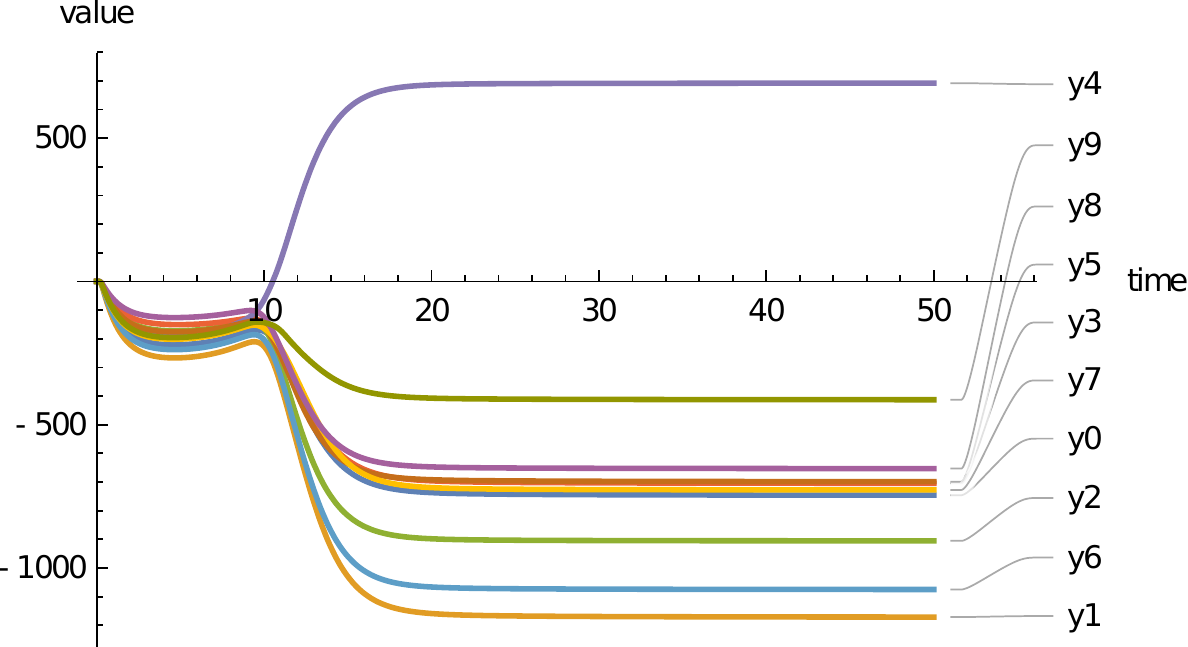}
    \caption{CRN simulation results.}
  \end{subfigure}
  \caption{Simulation of the MNIST CRN (part b) on an input image (part a). The image is encoded via the input species concentrations: e.g., value $0.54$ of the $65$th pixel ($4$th row, $9$th column) is represented via initial concentrations of $x_{65}^+$ and $x_{65}^-$ species. Simulation shows concentrations of the output species over time, i.e., the difference between positive and negative output species ($y_i = y_i^+ - y_i^-$). The CRN classifies the input image as a number $4$ (as $y_4$ has the highest value).}
  \label{fig:simulations}
\end{figure*}

\textbf{MNIST}.
We train a binary-weight neural network with $2$ hidden layers and $512$ units per each hidden layer,
where hyperparameter values are obtained via a random search.
We downscale MNIST images from original $28 \times 28$ resolution to $14 \times 14$;
thus the neural network has $14^2$ input units (one per pixel).
We use $10$ output units (for digits $0$ to $9$), and train the neural network to maximize the output unit corresponding to the correct digit.
Our trained model achieves accuracy of $97.76$\% on the test set.
Note that we did not focus on achieving high accuracy;
BinaryConnect in the original paper achieves accuracy of over $98$\%,
but uses larger networks ($3$ hidden layers with $1024$ units each). %
We translate the network to an equivalent CRN consisting of $4488$ chemical reactions (unoptimized compilation), and $1024$ chemical reactions (optimized compilation).
The CRN consists of $2 \cdot 14^2$ input species (two species per input unit encoding positive and negative parts),
and similarly $2 \cdot 10$ output species.
We simulate the CRN on $100$ randomly chosen examples from the test set, and confirm that the output matches that of the neural network in all of the cases.
Figure~\ref{fig:simulations} shows simulation results on an example input image (unoptimized CRN).
In regard to the dynamics of convergence, note the complex non-monotonic behavior resulting from the mass-action kinetics.

\begin{figure}[!t]
  \centering
  \includegraphics[scale=0.55]{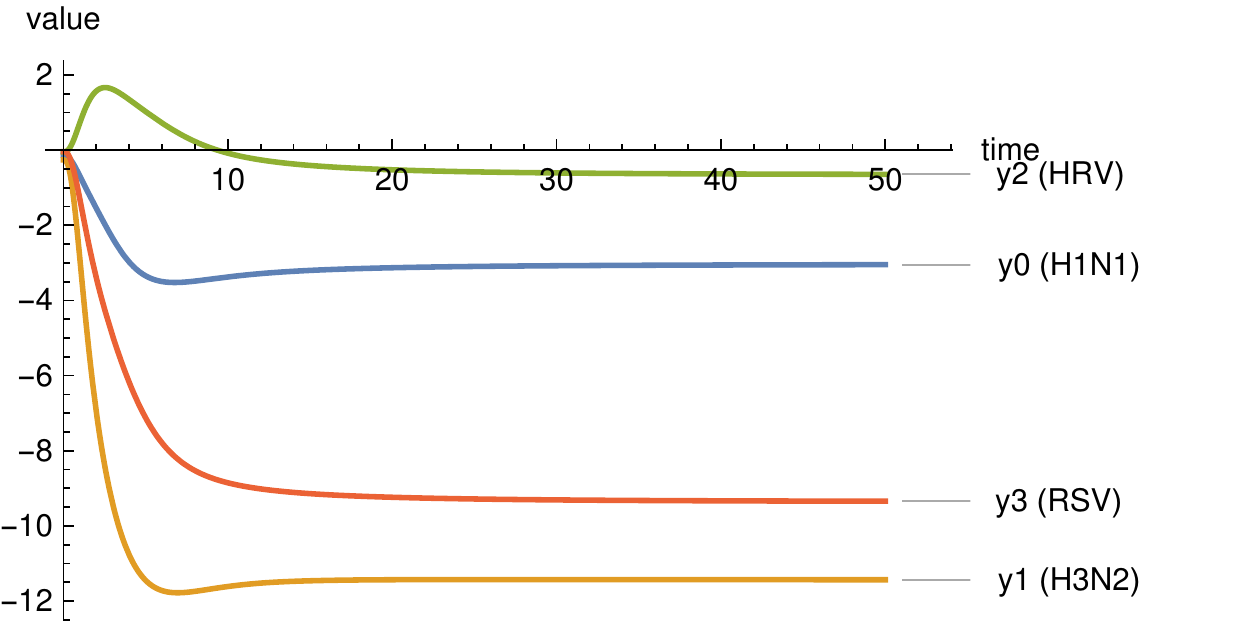}
  \caption{Simulation of Virus Infection CRN on an example (misclassified example). Example is classified as HRV virus, while the true label is H1N1.}
  \label{fig:disease-simulation}
\end{figure}

\textbf{Virus Infection}. We train a neural network with $1$ hidden layer with $32$ units, $10$ input units capturing expression of different genes, and $4$ output units classifying between virus infections.
We achieve test set accuracy of $95.20$\%.
The equivalent CRN consists of $148$ chemical reactions (unoptimized compilation), and $32$ chemical reactions (optimized compilation).
We simulate the CRN on $100$ randomly chosen examples from the test set and confirm that the output matches that of the neural network in all cases.
Figure~\ref{fig:disease-simulation} shows simulation of the unoptimized CRN on an example that is misclassified.

The performed experiments empirically confirm the soundness of our compilation technique.

\Subsection{Training Specifics}
We use the published implementation of BinaryConnect networks~\cite{courbariaux2015binaryconnect},
and follow the training procedure outlined in Section~\ref{sec:binaryconnect}.
We do not use Batch Normalization~\cite{ioffe2015batch}
because it would incur multiplication and division operations at the inference stage (training stage is not a problem) that would be hard to efficiently implement in CRNs.
Instead, we rely on Dropout~\cite{srivastava2014dropout} (stochastically dropping out units in a neural network during training) as a regularization technique.
In all our experiments we use the square hinge loss (as in BinaryConnect) with ADAM optimizer.

We train on MNIST dataset for $250$ epochs, measuring the validation accuracy at each epoch, and returning the model that achieves the best validation accuracy during training.
We train on IRIS dataset for $10,000$ epochs, and return the best performing epoch.
We train on the Virus Infection dataset for $200$ epochs, and return the model that achieved the best validation set accuracy.
We use an exponentially decaying learning rate. %

The rate constants of all reactions are set to $1$, and all chemical simulations are performed for $50$ arbitrary time units in the CRNSimulator package~\cite{CRNSimulatorPackage}.

\Section{Related Work}

It has long been observed that biological regulatory networks arguably behave in manner analogous to neural networks.
For example, both phosphorylation protein-protein interactions~\cite{hellingwerf1995signal,bray1995protein} and transcriptional networks~\cite{buchler2003schemes} can be viewed as performing neural network computation.

The challenge of implementing neural networks in chemical reaction networks also has a long history.
For example, Hjelmfelt et al~\cite{hjelmfelt1991chemical} propose a binary-valued chemical neuron, whose switch-like behavior relies on competition between excitation and inhibition. 
Recently, Moorman et al~\cite{moormandynamical2019} proposed an implementation of \Relu units based on a fast bimolecular sequestration reaction which competes with unimolecular production and degradation reactions. 
In contrast to the prior work, 
our implementation relies solely on the stoichiometric exchange of reactants for products, and is thus completely independent of the reaction rates.
Our CRN is also more compact, 
using only a single bimolecular reaction per neuron, with two species per every connection (without any additional species for the neuron itself). 
Finally, in contrast to the prior schemes, our CRN converges to a static rather than a dynamic equilibrium, which means that all reactions cease firing. 
This implies that our implementation does not waste energy to maintain state.

We use neural networks as a way to program chemistry.
The programming is done offline in the sense that neural networks are trained in silico.
However, there is a body of work on creating chemical systems that are capable of learning in chemistry~\cite{chiang2015reconfigurable,blount2017feedforward}.
Although these constructions are much more complex than ours, and arguably difficult to realize, they demonstrate the proof-of-principle that chemical interactions such as those within a single cell are capable of brain-like behavior.

Besides the above-mentioned theoretical work on chemical neural networks,
wet-lab demonstration of synthetic chemical neural computation argues that 
the theory is not vapid and that neural networks could be realized in chemistry.
A chemical linear classifier reading gene expression levels could perform basic disease diagnostics~ 
\cite{lopez2018molecular}.
Larger systems based on rate-dependent strand displacement cascades were used to implement Hopfield associative memory~\cite{qian2011neural},
and winner-take-all units to classify MNIST digits~\cite{cherry2018scaling}.

\Section{Conclusion}\label{sec:conclusion}
We demonstrate how BinaryConnect (weight $\pm 1$) neural networks could be implemented in chemistry using rate-independent chemical reaction networks.
As proof of principle, we demonstrate our scheme with numerical simulations of resulting CRNs classifying the MNIST and IRIS datasets.
We further simulate a CRN constructed from a ReLU network trained to classify a virus gene expression dataset.
Since this network relies on chemically available information for input, this example argues for the potential biological and medical utility of programming chemical computation via a translation from neural networks.

Performing the simulations with rate constants on the order of magnitude of those reported for DNA strand displacement ($10^3$ to $10^4$ M/s)~\cite{srinivas2017enzyme} results in roughly the conversion factor of 1 time unit = 100 to 1000 seconds of real time (assuming $10^{-6}$ M concentrations).
Although molecular systems may be orders of magnitude slower than electronics, the time scales are compatible with the intended applications such as interfacing with biological regulatory networks or disease marker detection. As a point of comparison, other implementations of chemical neural networks also take hours to complete~\cite{cherry2018scaling}.

The convergence time of the CRNs described here scales in a complicated way with many variables, from the depth of the implemented neural network to the choice of representing input values in dual-rail representation (an input $x = 3$ can be represented for example as $x^+ = 3$ and $x^- = 0$, or as $x^+ = 4$ and $x^- = 1$).
Thus an important area of future research is the formal proof of convergence properties of the system.
Empirically, we observed that depending on the network architecture, the 90\%-completion time may increase sublinearly, linearly, or superlinearly (in some contrived examples) with the number of layers.

Note that the concentrations of the positive ($S^+$) and negative ($S^-$) dual-rail species can become very large, despite the value they represent remaining small.
This can occur in special cases of the networks we construct.
This is undesired generally; in synthetic implementations, reactions can use additional ``fuel'' species in addition to the species designed in the formal CRN, so higher concentrations require more of such species.
However, there is an easy fix to this problem: inclusion of the reaction $S^+ + S^- \longrightarrow W$, where $W$ is some inert waste species.
This cancellation reaction leaves the signal value unchanged and reduces the concentrations of the two species.

Although in principle arbitrary CRNs can be implemented using DNA strand displacement reactions,
current laboratory demonstrations have been limited to small systems~\cite{srinivas2017enzyme},
and many challenges remain in constructing large CRNs in the laboratory.
Rate independent CRNs offer an attractive implementation target due to their absolute robustness to reaction rates.

Our construction is surprisingly compact in the sense that we use exactly one reaction per ReLU node.
This compactness argues that neural networks may be a fitting paradigm for programming chemical computation.
This invites the exchange of knowledge between the molecular programming and deep learning communities.

\section*{Acknowledgements}
We thank the anonymous reviewers for constructive comments and suggestions.  
This work was funded by NSF grants CCF-1718903 to SK and CCF-1901025 to DS.

\bibliographystyle{icml2020}
\bibliography{bib}

\end{document}